\pdfoutput=1
\documentclass[10pt,twocolumn,letterpaper]{article}

\usepackage[pagenumbers]{cvpr} 
\usepackage{amsfonts}
\usepackage{multirow}
\usepackage{bm}
\usepackage{physics}
\usepackage{mathtools}
\usepackage{float}
\usepackage[dvipsnames]{xcolor}

\makeatletter
\renewcommand{\paragraph}{%
  \@startsection{paragraph}{4}%
  {\z@}{0.4ex \@plus 1ex \@minus .2ex}{-1em}%
  {\normalfont\normalsize\bfseries}%
}
\makeatother
\usepackage[accsupp]{axessibility}

\newcommand{\RR}{\mathbb{R}}

\newcommand{\SE}{\mathrm{SE}\!\left(3\right)}
\newcommand{\SIM}{\mathrm{Sim}\!\left(3\right)}

\newcommand{\gD}{\mathfrak{D}}
\newcommand{\Se}{\Omega}
\newcommand{\Geo}{\mathcal{G}}
\newcommand{\KK}{\mathrm{K}}
\newcommand{\TT}{\mathrm{T}}
\newcommand{\II}{\mathrm{I}}
\newcommand{\PP}{P}
\newcommand{\MM}{\mathcal{M}}

\newcommand{\bu}{\bm{u}}
\newcommand{\bud}{\bm{\dot{u}}}

\newcommand{\bp}{\bm{p}}
\newcommand{\bpd}{\bm{\dot{p}}}

\newcommand{\bn}{\bm{n}}
\newcommand{\bx}{\bm{x}}

\newcommand{\Iref}{\II_{\text{ref}}}
\newcommand{\Isup}{\II_{\text{t}}}
\newcommand{\Trel}{\TT_{tr}}

\definecolor{cvprblue}{rgb}{0.21,0.49,0.74}
\usepackage[pagebackref,breaklinks,colorlinks,citecolor=cvprblue, linkcolor=magenta]{hyperref}

\def\paperID{9920} 
\def\confName{CVPR}
\def\confYear{2024}

\title{SuperPrimitive: Scene Reconstruction at a Primitive Level}
\author{Kirill Mazur, \quad Gwangbin Bae, \quad Andrew J. Davison \\
Dyson Robotics Lab, Imperial College London\\
{\tt\small \{k.mazur21, g.bae, a.davison\}@imperial.ac.uk}
}
\begin{document}
\maketitle

\begin{abstract}
    Joint camera pose and dense geometry estimation from a set of images or a monocular video remains a challenging problem due to its computational complexity and inherent visual ambiguities. Most dense incremental reconstruction systems operate directly on image pixels and solve for their 3D positions using multi-view geometry cues. Such pixel-level approaches suffer from ambiguities or violations of multi-view consistency (e.g. caused by textureless or specular surfaces). 

    We address this issue with a new image representation which we call a SuperPrimitive. SuperPrimitives are obtained by splitting images into semantically correlated local regions and enhancing them with estimated surface normal directions, both of which are predicted by state-of-the-art single image neural networks.
    This provides a local geometry estimate per SuperPrimitive, while their relative positions are adjusted based on multi-view observations.

    We demonstrate the versatility of our new representation by addressing three 3D reconstruction tasks: depth completion, few-view structure from motion, and monocular dense visual odometry. Project page: \hyperlink{https://makezur.github.io/SuperPrimitive/}{https://makezur.github.io/SuperPrimitive/}
    \vspace{-10pt}
\end{abstract}  
\section{Introduction}
\begin{figure}
  \includegraphics[width=1.0\columnwidth]{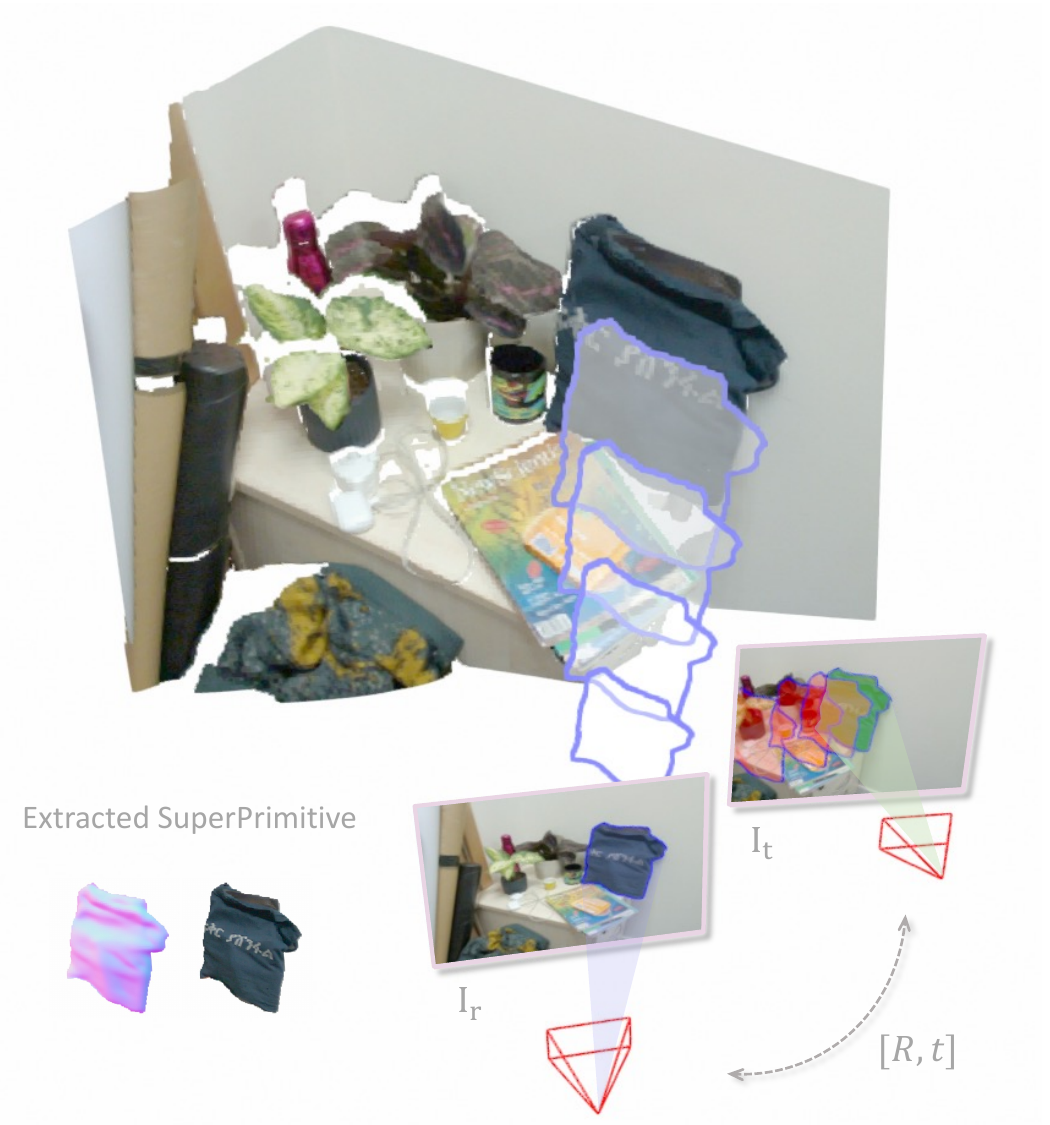}
  \caption{\textbf{Multi-View Geometry with SuperPrimitives.} 
  SuperPrimitives are extracted from an input frame by dividing it into image segments equipped with estimated surface normal directions (bottom-left). Each SuperPrimitive induces a dense reconstruction within the corresponding image segment up to \textit{a priori unknown} scale. Different possible reconstructions are shown in light blue. The scales are then jointly optimised together with a relative camera pose to fit multi-view photometric constraints (visualised in green and red). The resulting dense reconstruction of the reference frame is shown in the top.}
  \vspace{1mm} 
  \hrule
  \vspace{-1mm}
  \vspace{-16pt}
  \label{fig:teaser}
\end{figure}
Enriching monocular incremental reconstruction with prior world knowledge is essential for resolving visual ambiguities. This issue is particularly prevalent in scenarios with scarce data observations available: a notable example would be monocular visual SLAM, where images are being streamed from a camera into the system in real-time. 

When a monocular vision system encounters a new scene region, it must estimate the region's geometry based on a very limited number of observations. Without this, continuous camera motion tracking would not be possible. Once the scene region is thoroughly observed, the initial geometry estimate should be refined to better explain the multi-view information.

This naturally leads to a question: what sort of priors are effective in both providing reliable initial geometry estimates and supporting multi-view consistency? Geometric priors generally fall into one of two categories: local and global. Local priors, such as smoothness assumption~\cite{Newcombe:etal:ICCV2011} or surface normal regularisation~\cite{verbin:etal:CVPR2022}, impose additional constraints within a small neighbourhood. Global priors, on the other hand, aim to impose constraints on a larger scale, such as depth prediction~\cite{Eigen2014DepthMP, Bloesch:etal:CVPR2018}.

Our key observation is that some of the geometrical correlations are more reliable than the others, and therefore could be safely ``locked in'' together within local regions based on a single-view prediction. Points belonging to the same rigid body are strongly correlated, making it unnecessary to determine their depth independently. In contrast, distinct and unrelated objects can be placed arbitrarily in the scene. As the number of objects increases, learning a reliable global prior on their relative positions becomes an increasingly complex problem. 

In this work, we show that \textit{purely local but strong priors} are enough to achieve excellent performance across a variety of geometric vision tasks. For that purpose, we introduce a novel representation, \textit{SuperPrimitives}. A SuperPrimitive represents a local image segment, coupled with a dense shape estimate which is determined up to a scale factor. The scale factor can be further adjusted based on information observed from other views or additional measurements.

We show that SuperPrimitives can be efficiently constructed using a \textit{front-end} which consists of two single-image neural networks, extracting image segmentation and surface normal prediction. Effectively, our neural front-end predicts whether adjacent pixels belong to the same geometrical entity through image segmentation and estimates a surface normal at this point, thereby providing an infinitesimal geometry estimate. 

We delegate global, scene-level alignment to our streamlined multi-view, iterative, optimisation-based \textit{back-end}. The resulting front-end / back-end tandem combines the flexibility of multi-view based optimisation methods with the observation efficiency common in prior-driven systems.

Our new representation showcases its versatility in three key applications, where single-view ambiguity is resolved via additional measurements or viewpoints:
\begin{itemize}
\item Firstly, it adeptly handles zero-shot depth completion tasks in real-world scenarios, matching the performance of state-of-the-art methods tailored for depth completion;
\item Secondly, it facilitates joint pose and depth estimation using a limited set of unstructured images, surpassing its nearest competitor even in the absence of global priors;
\item Thirdly, our method outperforms previous monocular visual odometry systems on the challenging TUM dataset, and exhibits robustness across various domains.
\end{itemize}

\vspace{-4pt}
\section{Related Work}
\paragraph{Monocular Reconstruction.}
Both offline monocular reconstruction systems, such as COLMAP~\cite{schoenberger:etal:CVPR16} and \cite{FurukawaCSS10}, or online systems, such as MonoSLAM~\cite{Davison:etal:PAMI2007} and DSO~\cite{Klein:Murray:ISMAR2007} track, filter, and reconstruct only well-constrained points with high and reliable photometric information. This often involves fine-grained and sophisticated point management to reliably resolve visual ambiguities and filter unreliable visual observations. DTAM~\cite{Newcombe:etal:ICCV2011} demonstrated feasibility of incremental dense reconstruction in the monocular scenario. DTAM employed a hand-crafted local smoothness prior to handle regions with poor texture. Subsequently, other local priors~\cite{Weerasekera:etal:ICRA2017, Niemeyer:etal:CVPR2021} have been extensively explored to regularise multi-view geometry estimation problems.
\paragraph{Global Priors.}
We are interested in exploring the space of possible geometric priors from single-view networks. Depth prediction~\cite{Eigen:etal:NIPS2014, Ranftl:etal:2022, Ranftl:etal:ICCV21} is the most obvious choice but leads to a rigid per-image reconstruction which is difficult to feed into multi-view optimisation. In the recent years, deep-learning based approaches sought to replace explicit multi-view geometry estimating with learning-based methods~\cite{sun:etal:CVPR2021, bozic:etal:NIPS2021, sayed:etal:ECCV2022}. These methods, however, assume known poses and are, therefore, not suitable for joint pose and geometry estimation. Notably, CodeSLAM~\cite{Bloesch:etal:CVPR2018} introduced depth prediction conditioned on latent codes, which are then optimised to achieve cross-view consistency. However, CodeSLAM still struggles with out-of-domain data, as depth prediction networks are known to struggle with generalisation~\cite{Dijk2019HowDN}. 

\paragraph{Higher-Level Mapping.}
Introducing parametric primitives, such as lines~\cite{Vakhitov:eta:ECCV16, Vakhitov:etal:ICRA17}, planes~\cite{Gallup:etal:CVPR2010, Concha:etal:ICRA2014,  Kaess:ICRA2015, Liu:etal:CVPR18, Liu:etal:CVPR19, Shi:etal:ICCV23} or even high-order algebraic shapes~\cite{Cross:Zisserman:ICCV98, nicholson:etal:RAL2019, laidow:davison:3DV22} to better constrain multi-view geometry problems have been thoroughly explored over the last few decades. They all however use assumptions which may not often hold for all 3D scenes, especially for dense reconstruction. 

Besides parametric algebraic primitives object-level mapping has been explored in the last decade. SLAM++~\cite{Salas-Moreno:etal:CVPR2013} represented a map with a set of CAD models retrieved from an existing database and tracked camera position against these models. This method, however, could not be applied in a variety of settings due to the limited size of the CAD databases. Even objects of the same class can vary in their geometric appearance from instance to instance. This was later approached by~\cite{Sucar:etal:3DV2020, Fu:etal:RSS23}, who also represent their map via a set of objects, but learn a latent space of possible geometric variations per object class. 
Fusion++~\cite{McCormac:etal:3DV2018} extended this approach into a more versatile per object depth fusion using an RGB-D sensor and mask proposals from a pre-trained neural network. 
These methods rely on an RGB-D inputs and assume a pre-defined set of object classes, therefore lacking generality.
\vspace{-3pt}
\section{Method}

\begin{figure*}
  \includegraphics[width=1.0\linewidth]{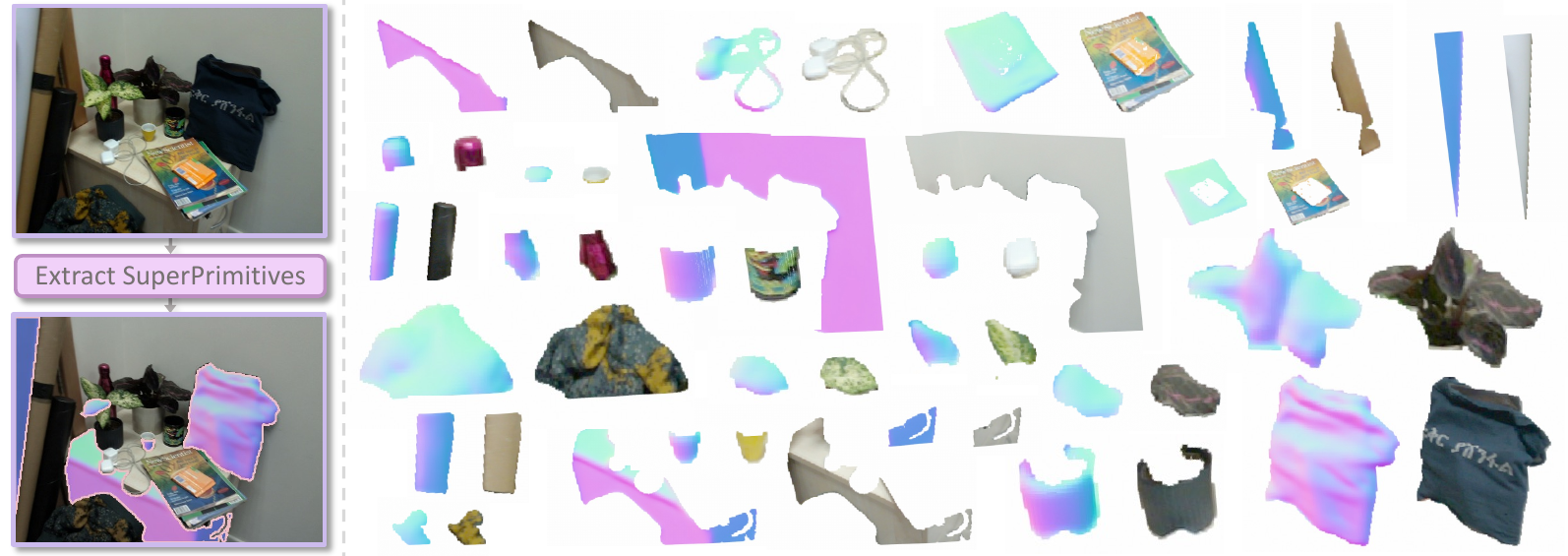}
  \caption{\textbf{SuperPrimitves Extraction.} \textbf{(left)} Our \textit{front-end processor} extracts SuperPrimitivies from an image by dividing it into a set of image regions with surface normal directions estimated for each image pixel within the segment. \textbf{(right)} Highlighted SuperPrimitives extracted from the image are visualised by showing their estimated normal and colour maps side by side. Note some of them are scaled either up or down for better viewing. While some of the SuperPrimitives are akin to object-level segmentation, the others tend to represent more low-level image segments.}
  \vspace{1mm} 
  \hrule
  \vspace{-1mm}
  \vspace{-14pt}
  \label{fig:frontend}
\end{figure*}

Firstly, we introduce the concept of a SuperPrimitive and explain how an image is processed into a set of its SuperPrimitives, referring to this part of our method as the \textit{front-end}. In the second part of this section, we describe how multi-view geometry problems can be reformulated at the level of SuperPrimitives instead of pixels. We refer to this stage of SuperPrimitive alignment as the \textit{back-end}.

The core of our method proposes splitting a given image into a set of (possibly overlapping) minimal segments, image regions which are likely to have strongly correlated geometry. We repurpose the recent state-of-the-art Segment Anything (SAM)~\cite{kirillov:etal:arxiv2023} model into predicting these minimal segments.

Our key idea is to estimate local geometry within each segment from a single view, while leaving the relative positioning of the segments to be estimated via multi-view photometric consistency optimisation. We refer to these geometrically enhanced minimal image segments as \textit{SuperPrimitives}, since they are inspired by both superpixels~\cite{Ren:etal:ICCV03} and geometric 3D primitives.  

For per-segment local geometry estimation we employ an off-the-shelf  surface normal prediction network~\cite{Bae:etal:ICCV2021}, to estimate infinitesimal geometry for each image pixel. The surface normals within an image segment can be used to estimate its depth via simple integration \textit{up to a scale factor}. We set these scale factors --- \textit{depth scales} --- to be optimisiable parameters, which are either optimised via multi-view cues or explicit depth measurements for the depth completion experiments. 

Thus, our method combines the strong priors provided by state-of-the-art neural networks in the front-end with the flexibility and consistency offered by multi-view optimisation.
\vspace{-5pt}
\subsection{Conventions}
Unless stated otherwise, we use lowercase letters for scalar values, bold letters for vectors, and uppercase letters for matrices. 
We consider images $\II \in \RR^{3 \times h \times w}$ of height $h$ and width $w$ captured by a camera with a known calibration matrix $\KK = \begin{psmallmatrix}
f_u & 0   & c_u \\
0   & f_v & c_v \\
0   & 0   &  1
\end{psmallmatrix} \in \RR^{3 \times 3}$. Image pixels are parameterised as $\bu = (u, v) \in \left[0, h-1\right] \times \left[0, w-1 \right]$.
Given a per pixel depth function $z(\bu) \colon \II \to \RR^{+}$, we define image's unprojection onto the 3D space to be $\pi^{-1}(\bu) =  z(\bu) \KK^{-1} \bud$, where dot is a homogenisation operator $ \bpd = (\bp, 1)$. Conversely, for a set of 3D points $\bx = (x, y, z) \in \RR^3$ we define the projection $\pi(\bx) = \gamma \left( \frac{1}{z} K \bx \right)$, where the $\gamma$ function drops the last coordinate $\gamma((x, y, z)^T) = (x, y)^T$.

We represent camera poses as matrices $\TT_{WC} \in \SE$ in the camera-to-world coordinates unless stated otherwise. During pose optimisation we store linearised pose increments $\mathbf{\xi} \in \mathfrak{se}(3)$ as Lie algebra elements. These increments are used then to update current camera pose estimates, see~\cite{Engel:etal:PAMI2017, Engel:etal:ECCV2014} for more detail.
\vspace{-3pt}
\subsection{SuperPrimitives}
We split the input image $\II$ into a set of possibly overlapping connected regions (segments), each of which is equipped with its local 3D geometry. More formally, a \textit{SuperPrimitive} $\PP = (\Se, \gD)$ is a connected image region $\Se \subseteq \II$ which is also equipped with an \textit{unscaled} depth map $\gD \colon \Se \to  [0, \infty )$ for each pixel within the region $\Se$. Here, unscaled means that each SuperPrimitive's depth $\gD$ differs from its ground truth depth $D$ by an \textit{a priori unknown} scalar. In other words, there exists a single scalar $s$ such that: 
{\setlength{\abovedisplayskip}{3pt} \setlength{\belowdisplayskip}{3pt}
\begin{align}
    D = s \cdot \gD  && \text{for all pixels p } \in \Se
\end{align}
}

We say that an image $\II$ is \textit{primified} into super-primitives $\mathcal{P}(\II) = \{ \PP_i = (\Se_i, \gD_i) \}$  if it has $n$, \textit{possibly overlapping} primitives that lie within the image $\II$, i.e  $\bigcup \Se_i \subseteq \II$.

For brevity we use the words \textit{primitive} and \textit{SuperPrimitive} interchangeably throughout the rest of the paper.
\paragraph{Representing Geometry with SuperPrimitives.}
\label{sec:geometry}
A set of primitives $\mathcal{P}$ itself is not enough to extract dense image geometry due to scale/depth ambiguity. Therefore, we introduce the concept of optimisable \textit{depth scales} which anchor the correct depth scale for each primitive. 

Given a scalar depth scale $s$ for a super-primitive $\PP$, one can infer its depth as $ D(p) = s \cdot \gD(p)$ for every point $p \in \Se$. In practice, we employ a log-depth representation to represent depth throughout the whole system. This means  that we store optimisable log-depth scales $\log s$ and unscaled log-depth $\log \gD$, and so depth inference reduces to a simple shift operation $\log D = \log s + \log \gD$.

In contrast to other dense methods, we estimate the dense geometry $\mathcal{G}$ of the image in the form of a point cloud rather than a depth image. This choice is driven by the fact that multiple primitives' supporting segments $\Se_i$ and $\Se_j$ might overlap, which would lead to different depth estimates within their intersection $\Se_i \cap \Se_j$.

Given a set of \textit{depth scaled primitives} $\mathbb{P} = \{ (s_i, P_i) \}$, the geometry $\Geo$ of the image $I$ can be estimated as follows:
{\setlength{\abovedisplayskip}{3pt} \setlength{\belowdisplayskip}{3pt}
\begin{align}
  \Geo \coloneqq \bigcup_i \, \pi^{-1}( s_i \cdot  \mathfrak{D}_i) 
\end{align}
}
To obtain estimated depth $\hat{D}$, we average depth values along the corresponding camera rays. 
\paragraph{Converting an Image into SuperPrimitives.}
Next we explain how SuperPrimitives are obtained in practice. An input image $I$ is first split into a set of minimal image segments $\Se_i$ produced by an image segmentation model. We use the Segment Anything~\cite{kirillov:etal:arxiv2023} model in this work, since it captures low-level semantic correlations highly accurately. Given the extracted segments, we independently estimate the local geometry $\gD_i$ within each segment by integrating surface normals predicted by another state-of-the-art neural network~\cite{Bae:etal:ICCV2021} within this region.

Note that this approach is premised on the assumption of geometry being continuous within each predicted image segment $\Se_i$, hence its unscaled depth values could be obtained by integrating its surface normals.
Although this is not guaranteed \textit{a priori}, we observed that a more fine-grained mask selection yields a compelling correlation with geometrical continuity. We discuss this in more detail in the following section.
\paragraph{Image Segment Retrieval.}
An ideal neural network which estimates image segments $\{\Se_i\}$ should predict regions of geometric continuity, where depth could be obtained via surface normal integration. However, to the best of our knowledge, such a network does not exist. Our approach aims to approximate this behaviour by utilising the Segment Anything model, coupled with a specialised mask selection process. Specifically, for each SAM query, we strive to select the smallest predicted mask surrounding that point. While such over-segmentation could potentially increase the dimensionality of the multi-view optimisation problem, under-segmentation might lead to incorrect geometry estimate within the primitive. Since we only adjust the scale of a primitive after the normal integration is completed, any incorrect geometry estimates within a primitive cannot be compensated at this stage.

The segmentation model employed in this work outputs three binary segmentation masks given a query point $q \in \II$. For each query $q$ we first filter predicted masks using the post-processing introduced in~\cite{kirillov:etal:arxiv2023}, such as stability filtering and Non-Maximum-Suppression (NMS). Even though our method allows redundant regions, we employed NMS to remove similar segments and save compute in the image alignment stage. With filtering done, the mask with the smallest area is selected. If the filtered set is empty, we discard the query point. 

We query the SAM backbone feature extractor once per image. Then we first sample $300$~query points randomly across the image, followed by the filtering discussed above. Then the image coverage mask is calculated and an additional $100$~mask query points are actively sampled in the uncovered regions.
\paragraph{Normal Integration.}
To estimate each segment's local geometry, we first pass the image through a surface normal estimation network. This network effectively predicts the derivative of desired depth values. Then, we ``integrate'' these surface normal vectors within each image segment to obtain its unscaled depth map.

For each image pixel $\bm{u} = (u, v)^T \in \II$ we estimate its surface normal vector $\bn = (n_x, n_y, n_z)^T$ with a pre-trained state-of-the-art CNN~\cite{Bae:etal:ICCV2021}. As was demonstrated in~\cite{Cao:etal:ECCV2022}, the log-depth $\tilde{z} = \log( z(\bm{u}))$ satisfies the following PDEs within a segment $\Se$:
{\setlength{\abovedisplayskip}{3pt} \setlength{\belowdisplayskip}{3pt}
\begin{align}
\tilde{n}_z \partial_u \tilde{z} + n_x = 0 && \text{and} && \tilde{n}_z \partial_v \tilde{z} + n_y = 0
\end{align}
}
 where $\tilde{n}_z = n_x(u - c_u) + n_y (v - c_v)  + n_z f$.
The depth values within the segment $\Se$ can therefore be obtained via minimising the following functional: 
{\setlength{\abovedisplayskip}{3pt} \setlength{\belowdisplayskip}{3pt}
\begin{align}
   \min_z \iint\limits_{\Se} (\tilde{n}_z \partial_u \tilde{z} + n_x)^2 + ( \tilde{n}_z \partial_u \tilde{z} + n_y)^2 \,\, du dv
\end{align}
}
Note that this leads to \textit{a family of solutions} $\tilde{z} + Const$ which differ by a shift constant. This is due to the fact that  only partial derivatives of $\tilde{z}$ are used in the functional. That means that the actual depth will be estimated up to scale, since $z = \exp(\tilde{z})$ by definition.

This ambiguity prompted us to introduce the notion of depth scales, to allow each segment to be adjusted towards its true depth. We implemented batched normal integration, efficiently solving all optimisation problems as a single sparse linear system using conjugate gradient method~\cite{Hestenes:etal:1952}. This implementation enables the integration of approximately $100$ to $300$ segments within a total time frame of $\sim\!100$ms.

\subsection{Primitive-based Image Alignment}
Now we explain how our new representation can be used as a building block for dense multi-view geometry and pose estimation, combining an optimisation-based mindset with learned single view priors. To make optimisation computationally tractable, we design our image alignment to \textit{fully abstract away any knowledge about the neural networks involved in the image primification stage}. Hence, even though there has been early evidence that SAM is independently capable of establishing (coarse) mask correspondences across neighbouring frames, we rely on photometric information only during the image alignment stage.
We believe that more semantic primitive association would be ``heavier'' and coarser since it would not provide per pixel correspondences, and would therefore \textit{not be a good fit to be used in a multi-view optimisation loop}. However, we speculate that such a method could be employed to further enhance our system, e.g. to improve occlusion handling, particularly as these models become more computationally affordable.
\vspace{-20pt}
\subsubsection{Two-view SfM on SuperPrimitives}
\vspace{-5pt}
\label{sec:sfm}
For the sake of clarity, we formulate our method in the simplest case of two frames observing the scene, although it could be trivially extended into a setting with a higher number of views. At the core of our Structure-from-Motion (SfM) approach lies a \textit{per-primtive photometric alignment}, which contrasts with widely accepted per-pixel photometric alignment~\cite{Newcombe:etal:ICCV2011, Engel:etal:PAMI2017} techniques. Informally speaking, our proposition is to treat each image primitive as a ``rigid'' piece, which is only allowed to be scaled up or down. This degree of freedom is due to the scale ambiguity present for each primitive. Thus, instead of estimating a depth value per pixel, we only estimate a depth-scale per primitive, which is illustrated in~\cref{fig:teaser}. This greatly reduces the dimensionality of the optimisation problem, especially in the case of an unknown relative pose.
Note that our method does not require the target image to be primified nor does it not require any pre-established correspondences (e.g. primitive to primitive).

We assume a primified reference image $\Iref$ with its set of primitives $\mathcal{P}(\II) = \{ (\Se_i, \gD_i) \}$. Given an \textit{unposed} target image $\Isup$, our goal is to \textit{jointly} estimate its relative pose $\Trel$ with respect to the reference image $\Iref$ as well as the dense geometry $\Geo_{\text{ref}} = \bigcup_i \, \pi^{-1}( s_i \cdot  \gD_i)$ of the reference frame.

While unscaled depths $\gD_i$ are given by a per-image pre-processor after the primification of the reference image, the set of depth scales $s_i$ are yet to be estimated. In our case, we jointly estimate both depth scales ${s_i}$ and a relative image pose $\Trel$ by solving a photometric consistency optimisation problem.
First, we warp each depth-scaled primitive $(s_i, P_i)$ from the reference frame $\Iref$ into the target frame $\Isup$:
{\setlength{\abovedisplayskip}{3pt} \setlength{\belowdisplayskip}{3pt}
\begin{align}
     \Hat{\PP_i} [\bu] = \pi \left(\Trel \pi^{-1} \left(\bu, s_i \cdot  \gD_i \right) \right)
\end{align}
}
Then a per-segment photometric residual for the primitive $\PP_i = (\Se_i, \gD_i)$ is defined by averaging all photometric reprojection $\ell^1$ errors across every pixel $\bu \in \Se_i$: 
\begin{align}
    r(\PP_i, s_i, \Isup, \Trel) = \frac{1}{|\Se_i|} \sum_{\bu \in \Se_i}   \norm{\Iref  (\bu) - \Isup (  \Hat{\PP_i} [\bu]  )}_1
\end{align}
The resulting photometric cost, aggregated across all depth-scaled primitives $(s_i, \PP_i) \in \mathcal{P}(\II)$, is:
\begin{align}
    E_{\text{photo}} = \frac{1}{|\mathcal{P}(\II)|} \sum_{P_i = (\Se_i, \gD)} r(P_i, s_i, \Isup, \Trel)
\end{align}
Note that we abstain from explicit occlusion handling in our photometric alignment, as our per-segment alignment is by design robust to pixel-level occlusion. We, however, expect our system could further be refined with explicit primitive correspondence checks.

Finally, to obtain the relative pose and depth scales, we minimise the photometric cost using the Adam~\cite{Kingma:Ba:ICLR2015} optimiser:
{\setlength{\abovedisplayskip}{3pt} \setlength{\belowdisplayskip}{3pt}
\begin{align}
 \label{eq:opt_problem}
  \{ s_i^{\text{opt}}, \Trel^{\text{opt}} \} = \operatorname{argmin}\limits_{s_i, \Trel} \, \, E_{\text{photo}}
\end{align}
}
Our approach trivially extends to multiple reference and target views via photometric cost summation, and we can jointly solve all depth scales and poses using the resulting cost function. 

\vspace{-7pt}
\subsubsection{Monocular Visual Odometry}
To demonstrate that our representation is suitable for \textit{simultaneous geometry and pose estimation}, we design a novel monocular visual odometry system which operates directly on primitives. Informally, we incrementally build a local 3D map out of primitives and then track new incoming frames against this map, also in a per-primitive manner. 

Visual Odometry (VO) consist of three components: initialisation, tracking, and mapping. Below, we cover each of these components. We interleave tracking and mapping in a single thread.

We adopt a keyframe-based approach~\cite{Klein:Murray:ISMAR2007}, which means that the 3D map is represented by a set of posed keyframes $\{ \II^i_{\text{kf}}, \TT^i_{\text{kf}}, \Geo^i \}$. Note that each keyframe also has its estimated dense geometry $\Geo^i$, which results in a three-dimensional local map of the scene. We map in a sliding window fashion, with a window size of $5$~keyframes. When the window is full, the earliest keyframe is popped from the window.
\paragraph{Initialisation.}
Initialisation of a VO system typically involves estimating the geometry of the first keyframe and its pose with respect to the subsequent few frames. Initialisation of a monocular incremental SfM or SLAM system is often hard, because one has to solve a chicken-and-egg problem: poses are required to reliably estimate geometry and vice versa. Therefore many VO / SLAM systems employ a wide range of heuristics in order to initialise their system.
However, using our new representation no special treatment is required in this case. In order to bootstrap our system, we simply create the two first keyframes and then employ the Structure-from-Motion method proposed in~\cref{sec:sfm}. We keep the pose of the first keyframe fixed to fix gauge freedom.
\paragraph{Tracking.} 
Given a map, the goal of tracking is to estimate the pose $\TT_{track} \in \SE$ of a new incoming frame $\II_{track}$. We adapt Lucas-Kanade~\cite{Lucas:Kanade:IJCAI1981} tracking onto our per-primitive formulation. At each step we track a new incoming frame against the latest keyframe in the odometry sliding window.

We solve the same photometric cost formulated in~\cref{eq:opt_problem}, but for the pose only. The latest keyframe serves as the reference frame $\Iref$ whereas the new incoming image is set to be the target frame $\Isup$. The depth scales of the latest keyframe have already been estimated at the mapping stage.

Since target images are not assumed to be primified, our tracking is essentially equivalent to the classical Lucas–Kanade method and hence could be implemented efficiently on modern hardware. Our proof-of-concept implementation does tracking at 2--3 FPS, but we expect high performance gains from employing standard machinery, such as Gauss-Newton optimisation~\cite{Leutenegger:etal:IJRR2014}.
\paragraph{Mapping.} 
The mapping stage ensures geometric consistency within the keyframe sliding window of size~$n$. Concretely, our mapping refines depth scales $\{s_j\}$ and a pose $\TT^i_{\text{kf}}$ for each keyframe $\II^i_{\text{kf}}$ where $0 \leq i \le n$.

For each keyframe, we define a set of target frames against which the photometric cost is being optimised with a \textit{connectivity function} $\MM(t) = \{\II_s, \TT_s\}$. In our implementation, $\MM(t)$ includes temporally neighbouring keyframes $\II^{t - 1}_{\text{kf}}$ and $\II^{t + 1}_{\text{kf}}$ (if they exist) used as supporting frames, as well as $4$ additional supplementary views, for which only the pose is being estimated. We employ these views to better constrain the geometry and observed that increasing the number of supplementary views does not induce high computational cost.

Thus, the mapping stage is done by solving a joint photometric cost for all keyframes:
{\small
\setlength{\abovedisplayskip}{3pt} \setlength{\belowdisplayskip}{3pt}
\begin{align}
    E_{\text{mapping}} = \sum_{t=0}^{n} \sum_{{\II_{s} \in \MM(t)}} \sum_{ 
 {P_i \in \mathcal{P}({\scriptscriptstyle \II^t_{kf}})}} r(P_i, s_i, \II_s, \TT^{-1}_{s} \TT_{kf})
\end{align}}

\section{Experiments}

\subsection{Sparse Depth Completion}
\begin{figure*}[h]
  \centering
  \includegraphics[width=1.0\linewidth]{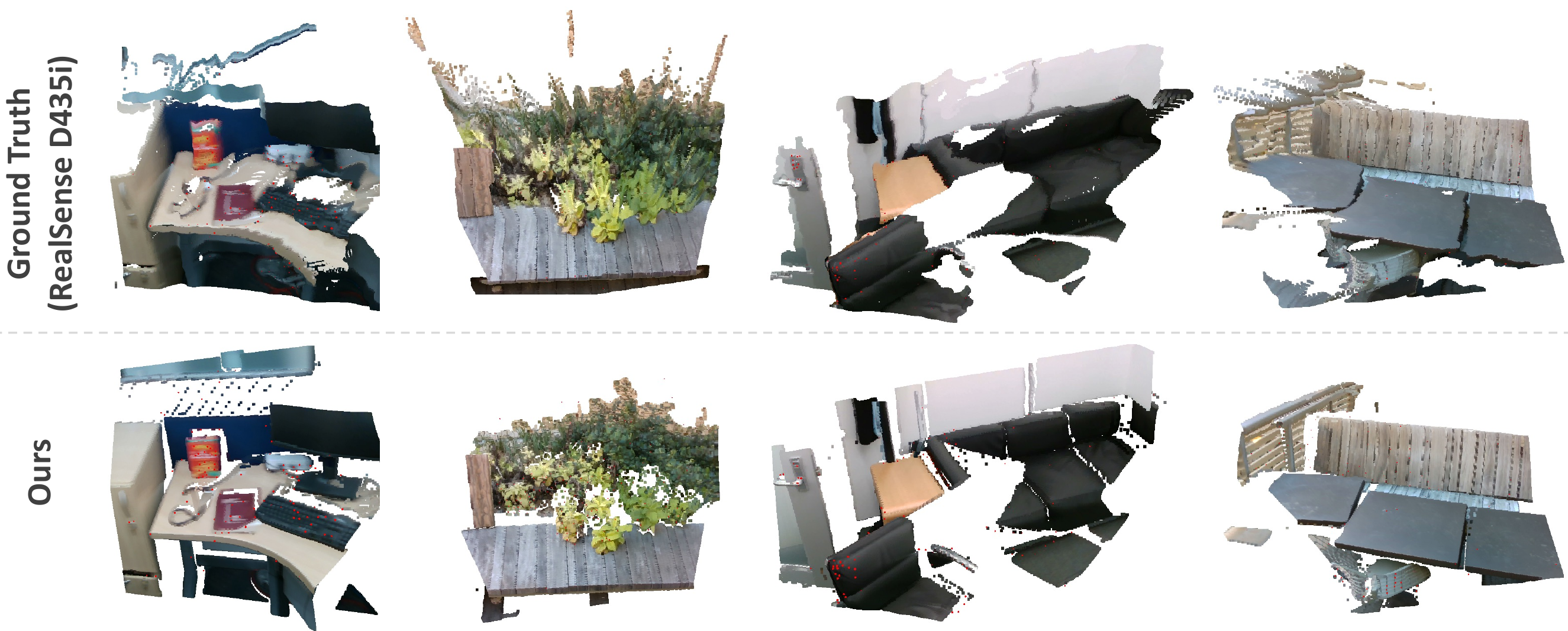}
  \caption{\textbf{Depth Completion on VOID.} We visualise the coloured unprojections of ground truth depth maps provided by a sensor (top row) and the geometry estimated by our method (bottom row). Sparse depth input points are visualised as red dots (electronic zoom-in recommended). Qualitatively, we achieve sharper geometry estimates than from a commodity depth sensor.}
  \vspace{-14pt}
  \label{fig:depth_completion}
\end{figure*}
\label{sec:depth_completion}
Our SuperPrimitives representation could be seamlessly applied to depth completion with no pre-training required, thereby solving it in a zero-shot manner. For each primitive $\PP_i \in \mathcal{P}(\II)$ we adjust its depth scale to minimise depth discrepancy $\| s_i \cdot \gD_i - \hat{D} \|$ with given ground truth sparse depth $\hat{D}$ across all valid depth points within the segment $\Se_i$. If a segment lacks valid depth measurements, it is discarded. For image regions not covered by any valid primitives, we generate a dense depth prediction by simply bilinearly interpolating depth values.

Many existing depth completion studies evaluate their methods in artificial scenarios, such as selecting input depth points randomly from a known ground truth. In contrast, our approach is tested using the real-world VOID benchmark~\cite{wong2020unsupervised}. The benchmark provides video sequences captured with a RealSense D435i camera together with sparse metric depth measurements acquired from an external visual-inertial SfM system. This setup exhibits noise and biases that might be present in the sparse depth inputs. These measurements are already provided by the dataset itself and are therefore shared across all the methods being compared. 

We compare our results with a recent state-of-the-art model~\cite{wofk2023videpth}, which significantly outperforms previous depth completion methods, particularly in zero-shot generalisation contexts. Other methods evaluated in~\cite{wofk2023videpth} are also reported in the table. Our depth completion method also operates in a zero-shot manner both task-wise and dataset-wise. That means our method was not trained for the depth completion task and neither of our surface normal prediction network nor segmentation model were trained on the VOID dataset. 

We focus on the most challenging ``150 points'' density setting, characterised by minimal sparse depth points per image. The depth is estimated in full $480 \times 640$ image resolution, following~\cite{wofk2023videpth}.  

In~\cref{tab:depth_comp} our method quantitatively performs on par with a recent state-of-the-art method, which is enabled by a monocular depth predictor pre-trained on a vast mixture of datasets~\cite{Ranftl:etal:2022, Ranftl:etal:ICCV21} and fine-tined on the VOID train set. In the zero-shot setting, we outperform VI-Depth~\cite{wofk2023videpth}, which uses a DPT-Hybrid pre-trained backbone, on three out of four metrics. Note that no training was done the for depth completion task for our method. 
Compared to the ground truth depth maps obtained via a noisy sensor, our predictions show (\cref{fig:depth_completion}) sharper object boundaries and are better at preserving straight lines and perpendicular structures.

\begin{table}
  \centering
  \scalebox{0.8}{
    \begin{tabular}{|@{\hspace{1mm}}c|@{\hspace{2mm}}c@{\hspace{2mm}}c@{\hspace{2mm}}c@{\hspace{2mm}}c@{\hspace{1mm}}|}
    \hline
    Method     & MAE     & RMSE   &  iMAE           & iRMSE       \\
    \hline
    VOICED~\cite{wong2020unsupervised}   & 174.04  & 253.14            &  87.39         &  126.3     \\
    NLSPN~\cite{park2020non}    & -       & -                 &  143           &  238.10     \\
    ScaffNet~\cite{wong2021learning} & 150.65  & 255.08            & 80.79          &  133.33     \\
    KBNet~\cite{wong2021unsupervised}    & 131.54  & 263.54            & 66.84          & 128.33      \\
    MonDi~\cite{Liu2022MonitoredDF}    & \underline{104.96} & 225.60           & 48.44          & 96.79      \\
    VI-Depth (DPT-Hybrid + NYUv2)  & - & -              & 55.90           & 85.20       \\
    \textbf{Ours}  & 109.0  & 204.15                  & \underline{47.32}         & \underline{83.40}  \\
    \hline
    VI-Depth${}^{*}$ (MiDaS)~\cite{wofk2023videpth} & 113.27 & \underline{193.38}   & 53.86           & 84.82      \\
    VI-Depth${}^{*}$ (DPT-Hybrid)~\cite{wofk2023videpth}  & \textbf{97.03} &\textbf{167.82} & \textbf{46.62}     & \textbf{74.67}      \\
    \hline
    \end{tabular}
  }
  \caption{\textbf{Depth Completion on VOID.} We report four most widely used metrics for depth completion on the VOID benchmark. The methods which did not use the VOID dataset for training are in the top section. Methods trained on VOID are marked with an asterisk${}^{*}$. Our method demonstrates superior performance on three out of the four metrics within the zero-shot group. It is second-best on two out of the four metrics overall.}
     \vspace{1mm} 
     \hrule
    \vspace{-1mm}
    \vspace{-15pt}
   \label{tab:depth_comp}
\end{table}

\subsection{Few-View Structure-from-Motion}
\begin{figure}
  \centering
  \includegraphics[width=1.0\columnwidth]{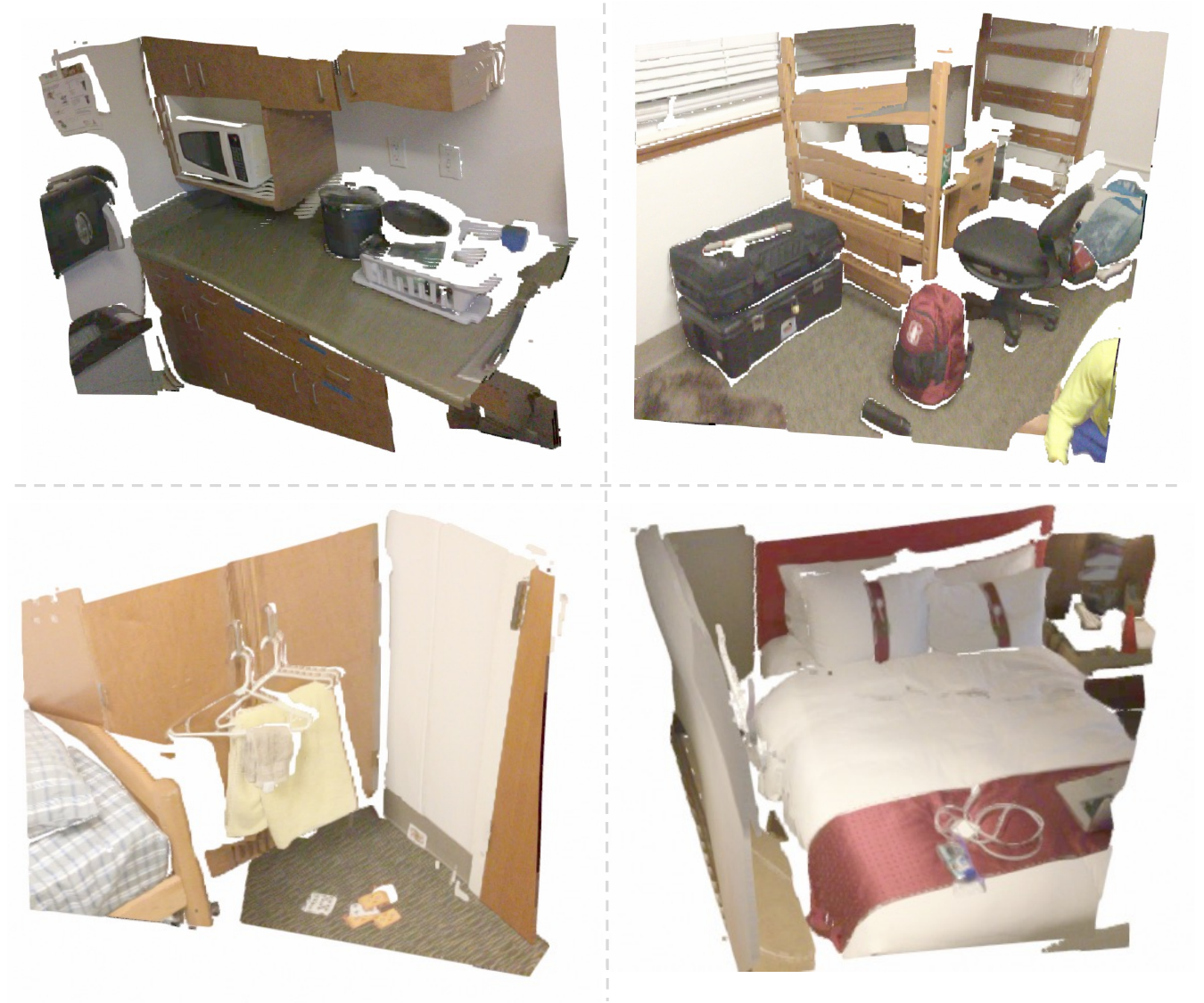}
  \caption{\textbf{3-View SfM on ScanNet.} We provide the visualisations of unprojected reference frame depth maps predicted by our method for few-view SfM using one reference and $2$ supplementary views. Note that we used surface normal prediction network which was only pretrained on HyperSim~\cite{roberts:2021} for this experiment.}
  \vspace{1mm} 
  \hrule
  \vspace{-1mm}
  \vspace{-15pt}
  \label{fig:sfm}
\end{figure}

\begin{figure*}[t]
  \centering
  \includegraphics[width=1.0\linewidth]{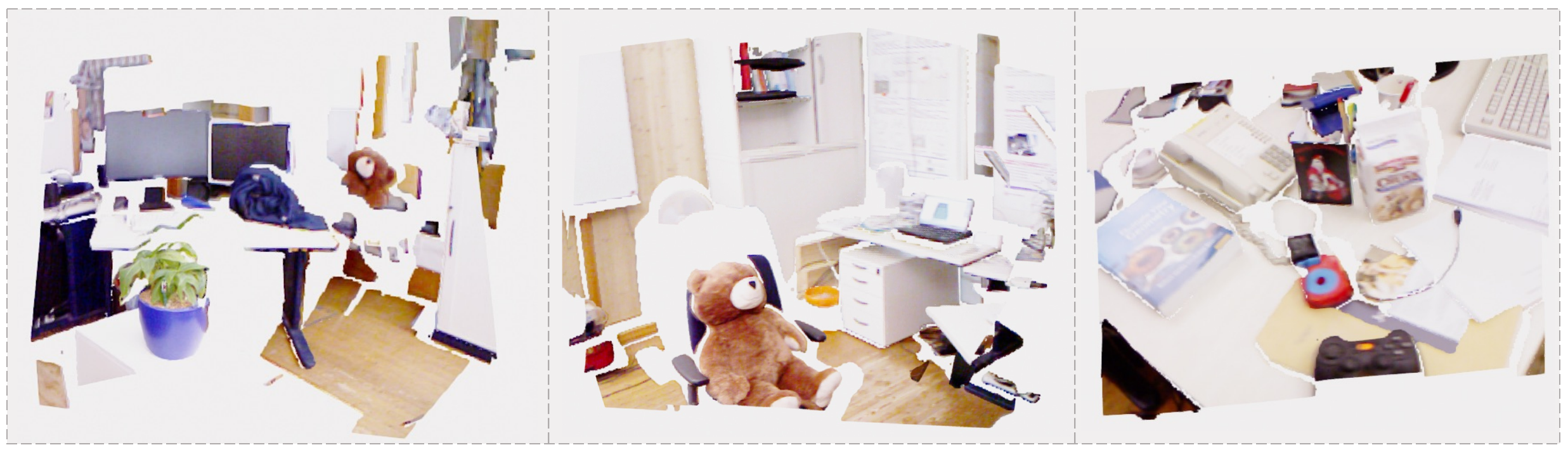}
  \caption{\textbf{TUM Reconstruction Results.} Examples of reconstructions produced by our monocular VO system on the TUM dataset. Each image shows a coloured point cloud of the geometry estimated on an odometry keyframe.}
  \vspace{-14pt}
  \label{fig:tum}
\end{figure*}

Contrasted with depth completion, where global geometry structure is roughly given, we also test our method on few view Structure-from-Motion. Given a set of \textit{unposed} images $\{ \Iref, \II_{s}^0, \ldots,  \II_{s}^{(n -1)} \}$ captured within a small timeframe, the goal is to estimate the depth of the reference image $\Iref$. In that setup, the set of unposed supplementary views $\{ \II_{s}^0, \ldots,  \II_{s}^{(n -1)} \}$ will be leveraged for multi-view geometry estimation. 

We choose the test set of the ScanNet dataset for this evaluation. In each test sequence we select every $200$-th frame to be the reference frame $\Iref$. Supporting frames are gathered from the neighbouring frames. Then we discard the frame sets with not enough motion to remove mostly static video clips, where SfM could not be performed. This resulted in $\sim$$500$ reference frames.
Note that in these experiments we use a surface normal neural network pre-trained only on the synthetic HyperSim dataset~\cite{roberts:2021}.

Since the multi-view depth estimation problem has scale ambiguity, we employ  median-scaling to align estimated depth to the metric scale for evaluation. We report iMAE and iRMSE in~\cref{fig:geomcomparison} for our method with varying number of supplementary views. We compare our depth estimation quality against the method closest to ours, DeepV2D~\cite{Teed:Deng:ICLR20}, which can also estimate depth together with supplementary frames poses. We demonstrate that our geometry quickly saturates after observing as little as 2 supporting views. Unlike DeepV2D we do not use any external tracking or initial relative pose estimation, yet still our method demonstrates consistent improvement over DeepV2D. Additionally, our method does not have any global prior on relative object positions, unlike DeepV2D.
Our approach therefore could be used as an VO / SLAM initialisation mechanism, estimating joint relative poses and geometry.
\begin{figure}
  \centering
  \includegraphics[width=\columnwidth]{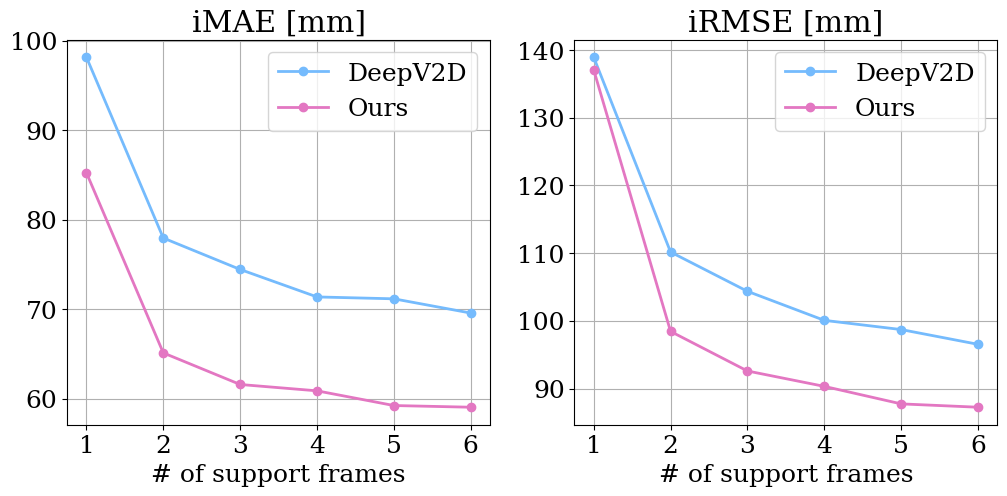}
  \caption{\textbf{Few-View SfM Depth Estimation Quality.} We evaluate the quality of our depth estimation method from an unstructured set of images on the ScanNet dataset. On the $x$-axis the number of supporting views is shown, while on the $y$-axis the corresponding depth reconstruction quality metric is reported. We show that the quality of our depth estimation quickly saturates and consistently outperforms its closest competitor, DeepV2D.}
  \vspace{1mm} 
   \hrule
 \vspace{-1mm}
  \vspace{-15pt}
  \label{fig:geomcomparison}
\end{figure}
\vspace{-3pt}
\subsection{Monocular Visual Odometry}
Monocular Visual Odometry requires both accurate pose and geometry estimation to successfully track camera motion across long trajectories. Minor pose estimation inaccuracies can accumulate over time resulting in what is called drift. With geometry being incorrectly estimated, accurate pose tracking becomes impossible, and vice versa. Our SuperPrimitive representation allows estimating both pose and geometry, enabling us to build a simple monocular VO system which performs better even in hard conditions.

We evaulate our monocular odometry on the TUM RGB-D~\cite{Sturm:etal:IROS2012} dataset. The dataset was captured with a handheld camera in indoor scenes. It is renowned for being incredibly challenging (especially for dense reconstruction systems), due to motion blur, rolling shutter artefacts, and abundance of pure rotational motion. We show that thanks to the strong priors encapsulated in our SuperPrimitives, our simple monocular odometry system can handle the TUM dataset without any special treatment (e.g. our method does not involve any special motion blur handling). 

We compare against other VO systems that do not have global bundle adjustment, following the protocol of~\cite{Dexheimer_2023_CVPR} and evaluate on 8~sequences from the Frieburg~1 split. We use only RGB images as the input to our system and downsample them to $120 \times 160$ for efficiency purposes. 
Since our method is purely monocular, the estimated trajectory lacks global scale and we first use $\SIM$ alignment to the ground truth scale, following standard practice~\cite{ORBSLAM3_TRO}. \cref{tab:tum_ate} shows that, despite the simplicity of our odometry system, it outperforms all other methods in terms of Average Trajectory Error (ATE)~\cite{Sturm:etal:IROS2012}, averaged across all trajectories. Additionally, our VO is either the best or second best on five out of eight sequences. Besides quantitative evaluation, we also demonstrate reconstruction results in~\cref{fig:tum}.

\begin{table}
  \centering
  \scalebox{0.9}{
    \begin{tabular}{|@{\hspace{1mm}}c|@{\hspace{2mm}}c@{\hspace{2mm}}c@{\hspace{2mm}}c@{\hspace{2mm}}c|@{\hspace{2mm}}c@{\hspace{1mm}}|}
    \hline
    \multirow{2}{*}{Seq.} & TartanVO & DeepV2D & DeepFactors & DepthCov & Ours \\
    & \cite{wang:etal:corl2020} & \cite{Teed:Deng:ICLR20} & \cite{Czarnowski:etal:RAL2020} & \cite{Dexheimer_2023_CVPR} & \\ 
    \hline
    360   & 0.178 & 0.243 & \underline{0.159} & \textbf{0.128} & 0.173            \\
    desk  & 0.125 & 0.166 & 0.170 &  \textbf{0.056} & \underline{0.085}           \\
    desk2 & 0.122 & 0.379 & 0.253 & \textbf{0.048} & \underline{0.108}            \\
    plant & 0.297 & \underline{0.203} & 0.305 & 0.261 & \textbf{0.153}                        \\
    room  & 0.333 & \textbf{0.246} & 0.364 & \underline{0.257} & 0.363            \\
    rpy   & \underline{0.049} & 0.105 & \textbf{0.043} & 0.052 & 0.055            \\
    teddy & 0.339 & \underline{0.316} & 0.601 & 0.475 & \textbf{0.253}            \\
    xyz   & 0.062 & 0.064 & \textbf{0.035} & 0.056 & \underline{0.036}            \\
    \hline 
    mean & 0.188  & 0.215 & 0.241 & \underline{0.167} & \textbf{0.153}            \\
    \hline
    \end{tabular}
  }
  \caption{\textbf{Trajectory Estimation Error on TUM.} Average Trajectory Error (ATE) is compared against other monocular odometry systems on the TUM Frieburg 1 split. The best and second best results are highlighted in \textbf{bold} and \underline{underscored} correspondingly. Our method outperforms others in terms of the ATE averaged across all trajectories.}
       \vspace{1mm} 
     \hrule
    \vspace{-1mm}
  \vspace{-14pt}
  \label{tab:tum_ate}
\end{table}

\section{Conclusion}
We presented a new representation, SuperPrimitive, which demonstrates how recent advances in building strong single image priors could be incorporated into pose and dense geometry estimation problems. We show that incorporating these priors alleviates the need for sophisticated hand-crafted heuristics and paves the way into monocular reconstruction with relative ease.
\vspace{7pt}

\paragraph{Acknowledgments.}
Research presented in this paper was supported by Dyson Technology Ltd. The authors would like to thank Hidenobu Matsuki, Eric Dexheimer, Riku Murai, and other members of the Dyson Robotics Lab for countless fruitful discussions. 
\maketitlesupplementary

\section{Ablation Study}
\subsection{Surface Normal vs Depth Prior}
While a monocular depth prior has been widely used in 3D reconstruction~\cite{Bloesch:etal:CVPR2018, Czarnowski:etal:RAL2020}, in this work we use a surface normal prior instead. This choice is driven by the fact that surface normals have stronger generalisation abilities and capture better scene geometry. This observation is confirmed with an ablation study on the depth completion experiment described in ~\cref{sec:depth_completion}. 

In this ablation experiment, we replaced our unscaled depth estimates $\gD$ (obtained by integrating surface normals) with the outputs of DPT-Hybrid~\cite{Ranftl:etal:ICCV21}, a state-of-the-art depth estimator. All depth scales are then optimised in the same way as in the original method.

\begin{figure}[t]
  \centering
  \includegraphics[width=1.0\linewidth]{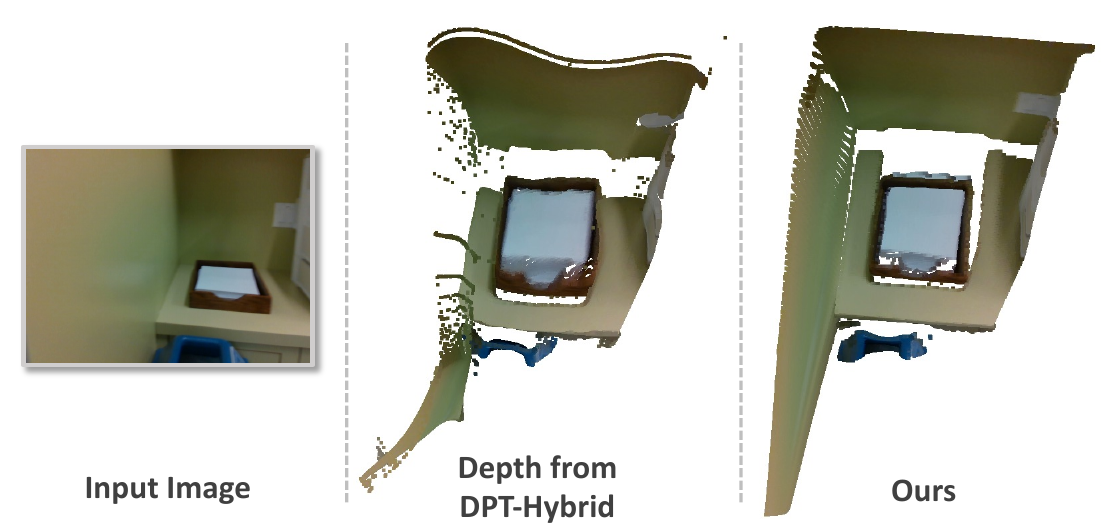}
  \caption{\textbf{Qualitative Ablation on VOID.} }
  \vspace{-14pt}
  \label{fig:normals_ablation}
\end{figure}

Our approach marginally outperforms its \mbox{DPT} counterpart on the VOID depth completion benchmark on all metrics (\cref{tab:depth_abl}). Qualitatively (\cref{fig:normals_ablation}), our method is better at preserving structural properties of the scene, such as walls.

\subsection{Surface Normal Quality Impact}

We investigate how the surface normal quality affects the performance of our method, especially for pose estimation. There are two ablation levels we performed in this case. The first one just assumes constant depth $z = const$ within each SuperPrimitive. This makes our method akin to Multiplane images (MPI) \cite{shade1998layered, zhou2018stereo}. To emulate planar but possibly slanted segments, we replaced surface normal vectors with their averaged value within each SuperPrimitive independently. Our full odometry system performs significantly better than its two ablated counterparts, see~\cref{tab:ate_abl}.

\begin{table}
  \centering
  \scalebox{0.8}{
    \begin{tabular}{|@{\hspace{1mm}}c|@{\hspace{2mm}}c@{\hspace{2mm}}|}
    \hline
    Method     & ATE    \\
    \hline
    Constant Depth  & 0.257  \\
    Constant Normals  & 0.208  \\
    \hline
    \textbf{Ours}  & \bf{0.153}    \\
    \hline
    \end{tabular}
  }
  \caption{\textbf{Ablation study on TUM.} We ablate our odometry by flattening out SuperPrimitives. See the main text for the details.}
   \label{tab:ate_abl}
\end{table}

\begin{table}
  \centering
  \scalebox{0.8}{
    \begin{tabular}{|@{\hspace{1mm}}c|@{\hspace{2mm}}c@{\hspace{2mm}}c@{\hspace{2mm}}c@{\hspace{2mm}}c@{\hspace{1mm}}|}
    \hline
    Method     & MAE     & RMSE   &  iMAE           & iRMSE       \\
    \hline
    Ours With Depth From DPT-Hybrid  & 128.03	& 213.32 &	57.86 &	91.82 \\
    \hline
    \textbf{Ours}  & \bf{109.0}  & \bf{204.15}  & \bf{47.32}  & \bf{83.40}  \\
    \hline
    \end{tabular}
  }
  \caption{\textbf{Ablation Study on VOID.} The quality of the unscaled depth obtained from method via surface normal integration is quantitatively compared against direct monocular depth estimation with DPT-Hybrid.}
    \vspace{-14pt}
   \label{tab:depth_abl}
\end{table}

\section{Implementation Details}
\paragraph{Segments Post Processing.} Segments $\Se_i$ extracted from the segmentation model may not be connected \textit{a priori}. We perform a simple mask connectivity check and split a primitive into two in the case of detected mask discontinuities.  

\paragraph{Depth scales.} Our depth scaling is implemented via storing a point $p_i$ within the segment $\Se_i$. We represent depth scales $s_i$ as the log-depth value at $p_i$. This point $p_i$ is the same as the query point provided to the segmentation model as an input. This depth scale parametrisation allows converting partially available depth maps into a set of depth scaled SuperPrimitives.

\subsection{Few-View Structure-From-Motion}
We initialised depth scales uniformly to $1.0$ for each primitive. All supplementary poses are initialised at identity. We solve SfM in a coarse-to-fine fashion to ensure the segments would not stuck in a local minima. 

A small penalty on depth scale was added with the weight $w = 1\mathrm{e}{-5}$ to constraint segments that may not have photometric information from other views.

\subsection{Depth Reinitialisation in MonoVO}
Given a new keyframe $\II^{\text{next}}_{\text{kf}}$ with an estimated pose $\TT^{\text{next}}_{\text{kf}}$, we scale depth for each new SuperPrimitve by using the geometry estimates of the previous keyframe $\II^{\text{prev}}_{\text{kf}}$. 

More precisely, we transform the point cloud $\Geo_{\text{prev}}$ of the previous keyframe into the coordinate system of the new keyframe and then render a partial depth map. Then, the depth scales $s_i$ of the new keyframe are estimated as in the depth completion experiments in~\cref{sec:depth_completion}.

\section{Experimental details}
\subsection{VOID Dataset}
For depth completion evaluation on the VOID dataset, we follow the protocol of~\cite{wofk2023videpth}. The ground truth depth is considered to be valid between $0.2$ and $5.0$ meters. The test set consists of $800$~images. The dataset also provides sparse depth measurements obtained by an external visual-inertial odometry. We choose the setting with least sparse depth measurements available, ``150~points'' --- where the SLAM system was configured to estimate depth of around $150$~feature points (which constitutes $0.05\%$ of the full image size). 

\begin{table}

\centering

\begin{tabular}{|c|c|c|}
   \hline
    Metric & Units & Definition \\ \hline
    MAE & {\it mm} &$\frac{1}{|\Omega|} \sum_{\bu \in\Omega} |\hat z(\bu) - z_{\text{gt}}(\bu)|$ \\
    RMSE & {\it mm} & $\big(\frac{1}{|\Omega|}\sum_{\bu \in\Omega}|\hat z(\bu) - z_{\text{gt}}(\bu)|^2 \big)^{1/2}$ \\
    iMAE & {\it 1/km} & $\frac{1}{|\Omega|} \sum_{\bu \in\Omega} |1/ \hat z(\bu) - 1/z_{\text{gt}}(\bu)|$ \\
    iRMSE & {\it 1/km} & $\big(\frac{1}{|\Omega|}\sum_{\bu \in\Omega}|1 / \hat z(\bu) - 1/z_{\text{gt}}(\bu)|^2\big)^{1/2}$ \\ \hline
\end{tabular}
\caption{\textbf{Error metrics.}  The definition of the error metrics used in depth quality valuation. Here, $\hat z$ and $z_{gt}$ are predicted and ground truth depth values respectively.}

\vspace{-2em}
\label{tab:error_metrics}
\end{table}

\subsection{TUM RGB-D Dataset}
For evaluation of the estimated trajectory we used ATE RMSE metric~\cite{Sturm:etal:IROS2012}. 

\section{Hardware Details}
All of our experiments were conducted in the following hardware setup: Intel Core i7 3.60GHz processor, 32~GB RAM, and NVIDIA GeForce RTX 4090 with 24GB VRAM.  Our method is implemented in PyTorch~\cite{Paszke:etal:ANIPS2019} and CuPy~\cite{cupy_learningsys2017} libraries. 

{
    \small
    \bibliographystyle{ieeenat_fullname}
    \bibliography{robotvision}
}

\end{document}